\documentclass[runningheads]{llncs}
\usepackage{graphicx}
\usepackage{cite}
\usepackage{amsmath,amssymb,amsfonts}
\usepackage{algorithmic}
\usepackage{textcomp}
\usepackage{xcolor}
\usepackage{hyperref}
\usepackage{multirow}
\usepackage{fontawesome}
\usepackage[utf8]{inputenc}
\usepackage{algorithm}
\usepackage{array}
\usepackage{algorithmic}
\usepackage{booktabs}
\begin{document}
\title{Locating and Mitigating Gender Bias in Large Language Models}
%
%
\author{Yuchen Cai\inst{1,2} \and
Ding Cao\inst{1,2} \and
Rongxi Guo\inst{1,2}\and
Yaqin Wen\inst{1,2}\and \\
Guiquan Liu\inst{1,2\textsuperscript{(\faEnvelopeO)}}\and
Enhong Chen\inst{1,2}}

\institute{University of Science and Technology of China, Hefei, China \and
State Key Laboratory of Cognitive Intelligence, Hefei, China\\
\email{\{caiyuchen, caoding, guorongxi, wyq\_65\}@mail.ustc.edu.cn, \{gqliu,cheneh\}@ustc.edu.cn}}

\maketitle              
\begin{abstract}
Large language models(\textbf{LLM}) are pre-trained on extensive corpora to learn facts and human cognition which contain human preferences. However, this process can inadvertently lead to these models acquiring biases and stereotypes prevalent in society. Prior research has typically tackled the issue of bias through a one-dimensional perspective, concentrating either on locating or mitigating it. This limited perspective has created obstacles in facilitating research on bias to synergistically complement and progressively build upon one another. In this study, we integrate the processes of locating and mitigating bias within a unified framework. Initially, we use causal mediation analysis to trace the causal effects of different components' activation within a large language model. Building on this, we propose the LSDM (\textbf{L}east \textbf{S}quare \textbf{D}ebias \textbf{M}ethod), a knowledge-editing based method for mitigating gender bias in occupational pronouns, and compare it against two baselines on three gender bias datasets and seven knowledge competency test datasets. The experimental results indicate that the primary contributors to gender bias are the bottom MLP modules acting on the last token of occupational pronouns and the top attention module acting on the final word in the sentence. Furthermore, LSDM mitigates gender bias in the model more effectively than the other baselines, while fully preserving the model's capabilities in all other aspects.

\end{abstract}
\section{Introduction}

In recent years, sophisticated artificial intelligence models, notably exemplified by ChatGPT \cite{wang2021gpt,touvron2023llama,yang2023baichuan}, are specially designed to excel in comprehending complicated natural language and generating human-like text. However, as these models become increasingly integrated across various sectors \cite{gilson2023does,choi2023chatgpt}, the inherent biases within these systems has become a subject of growing concern.

Bias refers to the existence of consistent inaccuracies, misattributions, or erroneous perceptions leading to a preference for specific groups or concepts, the reinforcement of stereotypes, or the formation of flawed conclusions derived from established patterns, it now profoundly impact users and society at large, underscoring the urgent need for a comprehensive examination and mitigation of these issues \cite{sun2019mitigating,ferrara2023should}.

Current research on bias primarily concentrates on two aspects: identification and location of bias \cite{caliskan2017semantics,vig2020causal}, which aims to understand the origins and manifestations of biases; and mitigation of bias \cite{bolukbasi2016man,ziegler2019fine}, which seeks to reduce or eliminate inherent biases. Due to the complex nature and extensive scope of bias manifestations, contemporary studies often focus on specific aspects, such as locating gender bias \cite{vig2020causal} or mitigating racial bias \cite{guo2022auto}.

Caliskan et al. \cite{caliskan2017semantics} adopts the core concept of the Implicit Association Test (IAT), measures gender bias by assessing the strength of conceptual associations, using the Word Embedding Association Test (WEAT) to evaluate bias in word embeddings. Caliskan et al. confirm that biases identified through IAT tests are present in word embeddings. Garg et al. \cite{garg2018word}  show that biases within word embeddings serve as indicators of societal changes, including the fluctuation of female participation in professional fields. Extending this further, May et al. \cite{may2019measuring} developed the Sentence Encoder Association Test (SEAT) from WEAT, extending its ability to evaluate bias in sentences. Jesse Vig et al. \cite{vig2020causal} introduces a methodology for interpreting neural language models to analyze how gender bias effects are mediated via specific model components in Transformer-based language models.

Numerous methods aimed at mitigating model biases have been successively proposed in recent years. Zhao et al. \cite{zhao2018gender} proposed creating an augmented dataset to mitigate gender bias in word embedding. This method involves training on the combined dataset, which includes both the original and gender-swapped data. The augmented dataset is generated using gender-swapping techniques. As word embedding models constitute a fundamental component in many NLP systems, mitigating biases within embedding significantly contributes to reducing the biases that propagate to downstream tasks. Bolukbasi et al. \cite{bolukbasi2016man} define gender bias as the correlation between the magnitude of the projection onto the gender subspace of a word embedding representing a gender-neutral word and that word’s bias rating, as rated by crowd workers. Bolukbasi et al. attempted to remove gender bias from the gender subspace. Additionally, Reinforcement Learning with Human Feedback (RLHF) \cite{ziegler2019fine}, an emerging debias method, was developed to fine-tune large language models, such as ChatGPT, in order to reduce their biases and align them with human values. This approach involves collecting a dataset of human demonstrations, comparisons, and preferences to create a reward model guiding the fine-tuning process \cite{ramamurthy2022reinforcement,cohen2022dynamic}.
Moreover, various debias methods based on masked language models have proven effective \cite{guo2022auto,zmigrod2019counterfactual}. Zmigrod et al. \cite{zmigrod2019counterfactual} employs a counterfactual data augmentation strategy by reversing gender pronouns found in Wikipedia, to reduce gender bias within the model through continued pre-training. Building on this foundation, Webster et al. \cite{webster2020measuring} adopt a dropout regularization strategy, aiming to reduce gender bias by increasing the Dropout rate during the continued pre-training.

Knowledge editing \cite{yao2023editing}, as an emerging technology, allows for selective updates and adjustments to the model’s knowledge without the need for complete retraining. These methods seek to balance the demand for accurate, current information with the practical limitations of computational resources and time. Some editing techniques have demonstrated robust editing capabilities, managing to modify up to 1000 factual knowledge points swiftly without impairing the model's other functionalities \cite{meng2022mass}. Some studies have attempted to transfer knowledge editing methods to other fields and have achieved success.
Cheng et al. \cite{cheng2023can, cheng2023editing} delved into the feasibility of editing multimodal large language models. They constructed a benchmark for multimodal language model knowledge editing scenarios and compared the editing effects of existing mainstream editing methods on multimodal large language models. Gandikota et al. \cite{gandikota2023erasing} delved deeply into the technique of erasing specific concepts from the weights of diffusion models through knowledge editing. Within the realm of diffusion models designed for image generation from text, this method demonstrated remarkable effectiveness.

While some studies have focused on identifying the locations of bias \cite{caliskan2017semantics, garg2018word,may2019measuring,vig2020causal}, the mechanisms underlying the generation of gender bias have not been thoroughly investigated. On the other hand, most of the proposed methods to mitigate gender bias focus on word embedding \cite{bolukbasi2016man}, special processing of training data \cite{zmigrod2019counterfactual}, or through human feedback \cite{ziegler2019fine}. How to exploit the mechanisms that generate gender bias for targeted gender debias and how to adapt knowledge editing techniques to the domain of gender bias mitigation remain under-explored.

To address the aforementioned issues, this study begins by analyzing the impact of different model components on the generation of gender bias, and the flow of biased information. Inspired by Vig et al. \cite{vig2020causal} and Meng et al. \cite{meng2022locating}, we use causal mediation analysis to trace the causal effects of different components' activation within a large language model. By running the model three times under different configurations, we observe the locations and mechanisms through which the model generates gender bias. The experimental results reveal that the bottom MLP modules acting on the last token of occupational pronouns and the top attention modules processing the final word of the sentence play significant roles in influencing the generation of gender bias. To further validate the influence of the bottom MLP modules on the generation of gender bias, we have adapted a knowledge editing technique \cite{meng2022mass} to mitigation of gender bias within the model, called \textbf{LSDM} (\textbf{L}east \textbf{S}quare \textbf{D}ebias \textbf{M}ethod). \textbf{LSDM} modifies parameters by solving a matrix equation with constraint terms, enabling us to minimize interference with other aspects of the model while specifically mitigating gender bias associated with certain occupation words. \textbf{LSDM} overcomes the catastrophic forgetting problem that exists in all other debiasing methods and stands out by avoiding additional reinforcement learning or human annotations, basing its approach on causal trace conclusions rather than just black-box fine-tuning. Our main contributions are as follows:
\begin{itemize}
    \item[$\bullet$] We trace the causal effects of different components' activation within a large language model using causal mediation analysis to measure the magnitude of the impact of different components of the model on gender bias and reveal the flow process of biased information.
    \item[$\bullet$] We propose \textbf{L}east \textbf{S}quare \textbf{D}ebias \textbf{M}ethod to modify parameters to mitigate gender bias in models. This is a more interpretable debiasing algorithm. Results confirm that \textbf{LSDM} serves as an efficient debias method that overcomes the catastrophic forgetting problem that exists in all other debiasing methods.
    \item[$\bullet$]To our knowledge, this study represents the first endeavour to incorporate both the location and mitigation of gender bias into a unified framework.
    \item[$\bullet$]We are the first to transfer knowledge editing methods to the domain of debias and validate their feasibility, providing a viable solution for eliminating various biases present in large language models.
\end{itemize}

\section{Interventions on Activations for Tracing Bias Information Flow}
\subsection{Preliminaries}\label{Preliminaries}
Our primary focus lies on autoregressive, decoder-only language model structures denoted as $\mathcal{F}_{\theta}$, representing the predominant architecture found in contemporary large language models. These models operate by transforming the input sequence ${\boldsymbol{x}}$ into $t$ tokens $x_0, ..., x_t$. Subsequently, these tokens are fed through $L$ layers of Transformer decoders, ultimately generating probabilities for the next token $x_{t+1}$:
\begin{equation}\label{pmet.llms}
\begin{aligned}
  \mathcal{F}_{\theta}(x_0, ...,x_t) &=\text{softmax}\left(W_{\text{E}}\cdot\gamma\left(h_t^{L-1}+a_t^L+m_t^L\right)\right)\\&= \mathbb{P}\left(x_{t+1}|x_0, ...,x_t\right)
  \end{aligned}
\end{equation}

Here, $W_E$ and $\gamma$ represent the embedding matrix and layernorm, respectively. $a_z^L$ and $m_z^L$ denote the hidden states of the Multi-Head Self-Attention (MHSA) and Feed-Forward Network (MLP) at the $L$-th layer. The general forms of MHSA and MLP at the $l$-th layer and the $j$-th token $x_j^l$ are given as follows:
\begin{equation}\label{eq2}
\begin{aligned}
\footnotesize
  a_j^l &= W^l_{\text{MHSA}}\cdot\text{MHSA}^l\left(\gamma\left(h^{l-1}_1,  h^{l-1}_2,...,h^{l-1}_j\right)\right),\\
  m_j^l &=W_{proj}^l\cdot\sigma\left(W_{fc}^l\gamma\left(a_j^l+h_j^{l-1}\right)\right),\\
  h_j^l &= h_j^{l-1} + a_j^l + m_j^l
\end{aligned}
\end{equation}

\begin{figure*}[htb]
\centering
\includegraphics[width=\linewidth]{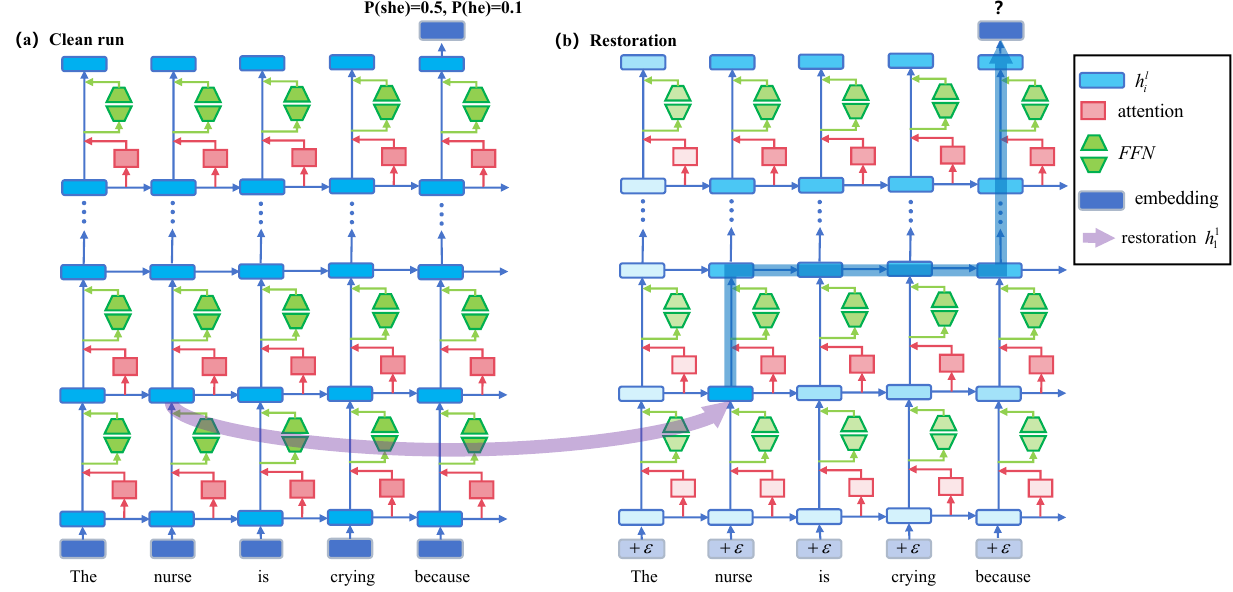}
\caption{\textbf{Causal Trace} computes the causal effect of neurons activation by running the network three times: (1)
once normally, (2) once where we corrupt the embedding, (3) once where we corrupt the embedding and then restore selected internal activation to their clean value. (a) The \textbf{clean run} procedure is depicted on the left side. (b) The \textbf{restoration} process is presented on the right side, measures the impact of $h_1^1$ among the generation of gender bias by corrupting in embedding and restoring the $h_1^1$ of the first layer.}
\label{fig1}
\end{figure*}
\begin{figure*}[htb]
\centering
\includegraphics[width=\linewidth]{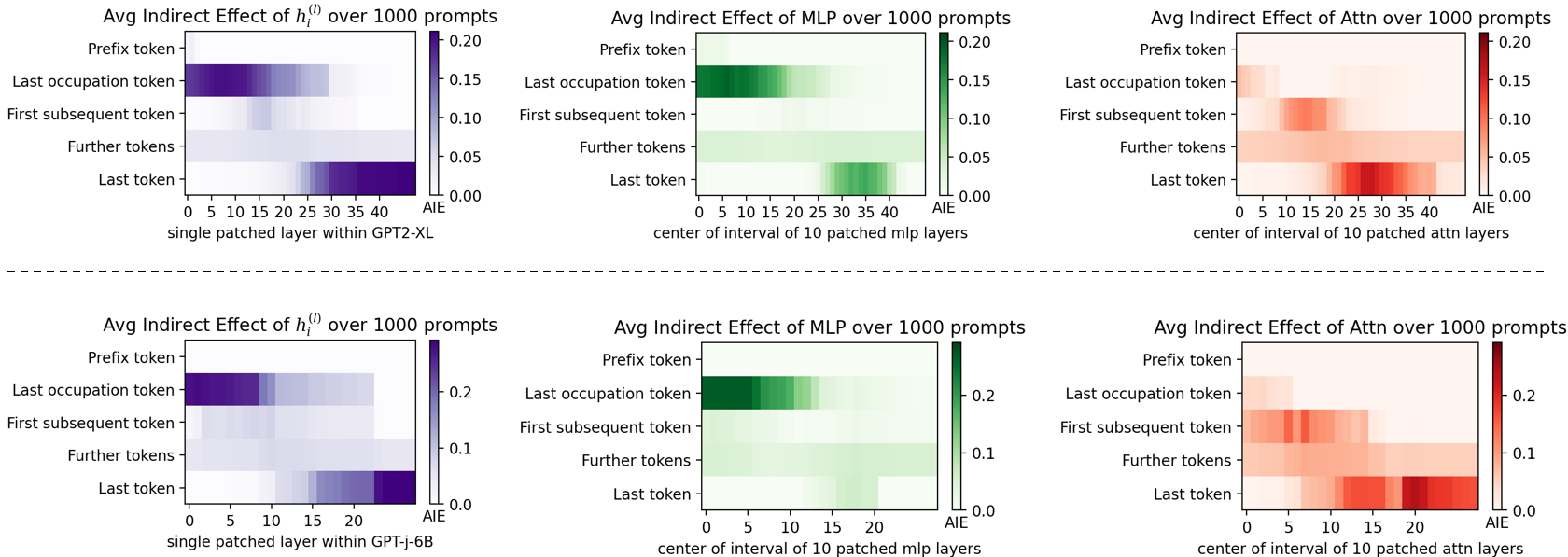}
\caption{The upper and lower parts correspond to the causal trace results of GPT2-XL and GPT-J-6B. The horizontal axis represents the different layers of the model, while the vertical axis signifies the meanings of various tokens within a sentence. The color of the graph indicates the magnitude of the corresponding AIE.
For the MLP and attention modules, considering the minimal influence that altering a single MLP or attention has on the model, we concurrently adjust the MLP (attention) values across a suite of 10 layers at once, centered on layer $l$, including the 5 layers above and 4 layers below it, as with Meng et al \cite{meng2022locating}.}
\label{fig2}
\end{figure*}

Here, $W^l_{\text{MHSA}}$ and $W_{proj}$ refer to the output weights of the MHSA and MLP at the $l$-th layer, respectively, while $\sigma$ denotes the non-linear activation function. Different LLMs frequently exhibit slight variations in implementing these transformations. Our goal is not to provide a full survey of these details but to capture essential terminology for our results.

\subsection{Causal Tracing of Gender Bias}\label{casual trac}

Based on Vig et al. \cite{vig2020causal} and Meng et al. \cite{meng2022locating}, we aimed to delve into the mechanisms generating gender bias and its storage locations. Figure \ref{fig1} illustrates a causal graph, detailing the components within a large language model used to predict the subsequent word and the workflow for causal tracing. We quantify the extent of gender bias within the model by calculating the absolute value of the disparity between the predicted probabilities for ``she" and ``he": $\mathbb{P}(gb) = \left|\mathbb{P}(he) - \mathbb{P}(she) \right|$. we assess each component contributions to $\mathbb{P}(gb)$ across three different runs:

\begin{itemize}
    \item[$\bullet$] In the \textbf{clean run}, we pass a sentence including occupational pronoun that may be associated with gender bias into model $G$ and collect different component's activation. Figure \ref{fig1} provides an example illustration with the sentence: ''The nurse is crying because \rule{0.5cm}{0.15mm}'', for which the model prefers to predict the next word as ``she'' rather than ``he''.
    
    \item[$\bullet$] In the \textbf{corrupted run}, the sentence undergoes obfuscation before the network's execution. Embeddings corresponding to all tokens within the sentence are corrupted as $h_{i*}^0:= h_{i}^0 + \epsilon$, where $\epsilon \sim N(0;\nu)$\footnote{We set $\nu$ to be three times larger than the empirical standard deviation of embeddings. Refer to Meng et al. \cite{meng2022locating} for specifics.}. This alteration could influence the model's prediction probabilities for gender-associated words.
    
    \item[$\bullet$] In the \textbf{restoration}, the causal impact of internal hidden states is examined by restoring these states to their values from the \textbf{clean run}. We intercept a hidden state, setting it to output the clean state $h_{i*}^0 := h_{i}^0$; subsequent computations proceed without additional intervention. Essentially, if a finite set of unaffected states can reliably predict the model's probabilities for the words ``she" and ``he", it indicates that they play a pivotal role in shaping the model's information processing.
\end{itemize}

Let {$\mathbb{P}(gb)$}, {$\mathbb{P}_{*}(gb)$}, {$\mathbb{P}_{*, clean}(gb)$} denote the probability of sentences under the clean, corrupted, and corrupted-with-restoration runs. We define \textbf{total effect} (TE) as: TE = $\mathbb{P}(gb)$ - $\mathbb{P}_{*}(gb)$, this shows the magnitude of the effect of a small perturbation on the gender bias of the model. \textbf{Indirect effect} (IE) evaluates the impact of restoring a specific $h_{i}^l$ of a token on the recovery of model gender bias. It is defined as follows: IE = $\mathbb{P}_{*, clean}(gb)$ - $\mathbb{P}_{*}(gb)$. By averaging over a sample of sentences, we can determine the Average Total Effect (ATE) and the Average Indirect Effect (AIE) for each hidden state variable.

\subsection{Causal Tracing Results}\label{causal}

We utilized GPT2-XL \cite{Radford2019LanguageMA} and GPT-J \cite{wang2021gpt} as our experimental models due to their range of layers, spanning the upper and lower bounds of current large-scale language models. Employing 1000 biased sentences containing occupational pronouns, we computed ATE and AIE. Our analysis involved altering the activation values of different components corresponding to each token in the sentence,  not only include hidden states $h_i^{l}$, but also MLP modules $m_i^l$ and attention modules $a_i^l$. 

\begin{figure*}[htb]
\centering
\includegraphics[width=\linewidth]{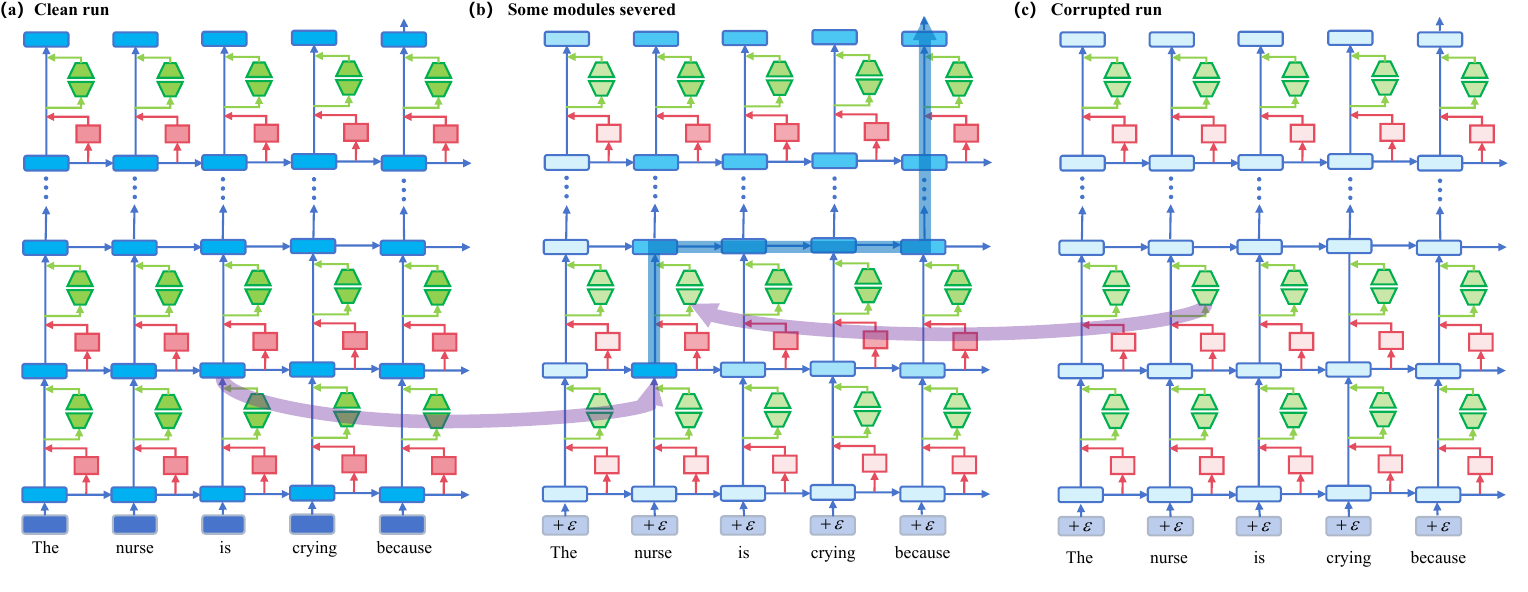}
\caption{To isolate the effects of MLP (attention) modules when measuring causal effects, the computation graph is modified.}
\label{fig3}
\end{figure*}

The computed ATE of 0.229 and 0.354 indicate perturbations in the behavior of the model due to corruption in word embedding. Figure \ref{fig2} illustrates the AIE of internal components of GPT-2 XL and GPT-J. Drawing from Figure \ref{fig2}, the following conclusion can be inferred:
\begin{itemize}
    \item[$\bullet$]The two higher portions of the AIE values in    
      $h_i^{l}$ corresponds to the higher AIE values in MLP and attention module respectively, with one located at the bottom layer of the last occupation token and the other at the top layer of the last token. This is understandable since, according to Eqn.\ref{eq2}, the value of $h_i^{l}$ is jointly constituted by $m_i^{l}$ and $a_i^{l}$.
    \item[$\bullet$]While high AIE values are observed in both MLP and attention modules, the higher AIE values in the attention module appear at the top layers near the model's output, do not constitute a novel insight. However, in the MLP module, the high AIE values emerge at an early stage — the bottom layer, and specifically in the middle part of the sentence — the last occupation pronoun token, marking a significant new discovery.
    \item[$\bullet$]Our experiments reveal notable differences between GPT2-XL and GPT-J: the significance of the bottom MLP modules at the last occupational token and the top attention modules of the final token is more accentuated in GPT-J. Conversely, GPT2-XL exhibits contributions from top MLP modules, such as layer 35. These distinctions indicate nuanced variations in the bias transfer processes among models with different parameters.
\end{itemize}

To gain deeper insight into the distinct role of MLP at the bottom layer, we analyzed indirect effects using a modified causal graph (Figure \ref{fig3}). Here are the steps of our approach: 
\begin{itemize}
    \item[$\bullet$]Initially, we collected the contribution of each MLP module within baselines with corrupted input embedding.
    \item[$\bullet$]Next, aiming to distinctly isolate the influence of MLP modules in quantifying causal effects, we modified the computation graph and substituted the computation of the MLP with MLP computations performed in the corrupted state (\textbf{restoration}), ensuring they remain unaffected the insertion of clean state for $h_{i}^{l}$. 

\begin{figure*}[htb]
\centering
\includegraphics[width=\linewidth]{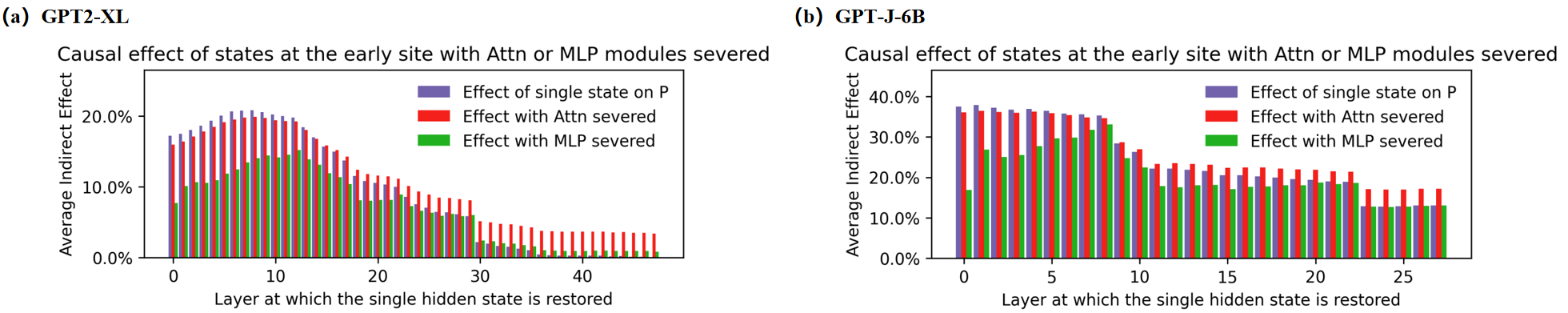}
\caption{Comparing AIE with and without isolating MLP (attention) modules.}
\label{fig4}
\end{figure*}

    \item[$\bullet$]We sought to clarify the contributions of the MLP module in the early stage by contrasting the AIE in the modified graph with those in the original graph.
\end{itemize}

In Figure \ref{fig4}, we observe the bottom layers lose their causal effect without the activity of MLP modules, while the top layer states’ effects depend little on the MLP activity. This result confirms an important role: the bottom MLP generates gender bias in calculations.

Those experiment results corroborate earlier investigations by Dai et al. \cite{Dai2021KnowledgeNI} and Meng et al. \cite{meng2022locating}, focusing on knowledge storage and transfer within language models. Their research underscored that a significant portion of a language model's knowledge resides within the MLP layer, characterized by extensive involvement, while attention serves to facilitate information exchange between tokens. Geva et al. \cite{Geva2022TransformerFL, Geva2023DissectingRO} conducted experiments supporting this hypothesis, noting alterations in the model's output upon disabling parts of the attention mechanism. They proposed that the representation at the subject position undergoes an enrichment process, driven primarily by the bottom MLP modules, aimed at encoding various subject-related attributes. This enriched information is then transmitted to facilitate prediction.

Drawing from these experiments and prior studies, we propose that \textbf{Gender bias information emerges due to the actions of MLP modules on specific occupational pronouns. Subsequently, this biased information is captured by the top attention modules, ultimately affecting the model's output.}

\section{Interventions on Weights for Mitigating Gender Bias}
\subsection{Associative Memory Foundation}
The concept of associative memory posits that any matrix $W$ can function as an associative memory \cite{meng2022locating}, storing a set of key-value pairs ${(k_i, v_i)}$ retrievable through matrix multiplication:

\begin{equation}
v_i \approx Wk_i,
\end{equation}

This principle stands as a foundational tenet in neural networks \cite{Kohonen1972CorrelationMM}. For instance, if the keys ${k_i}$ form a set of mutually orthogonal unit-norm vectors, a flawless memory can be constructed as $W = \sum_i v_i k_i^\top$. In relative terms, an $M \times N$ matrix can hold a maximum of $N$ associations.

In the preceding section, we outlined the bias generation mechanism, attributing bias generation to the bottom MLP modules within the model. This suggests that after enriching information within the bottom MLP modules, the biased vector is formed. Consequently, information from this biased vector is extracted by top attention modules, significantly affecting the model's output. When combined with the associative memory concept, it leads to an inference: if an unbiased vector $k$ exists, after sequential transformations through some associative memory matrices (i.e., the bottom MLP modules\footnote{MLP module contains two associative memory matrices. Refer to \ref{Preliminaries} for details.}), these matrices will progressively alter vector $k$ into a vector $v$ with gender bias. Considering the impact of mutiple associative memory matrices, we can express this transformation as:
\begin{equation}
\begin{aligned}
v_{bias} = W_{total}k.
\end{aligned}
\end{equation}

\subsection{Least Square Debias Methd}\label{lsdm}
Based on what was mentioned earlier, if we can consider the bottom MLP modules as associative memories, the certain vector $k$ representing the occupation-related word (such as ``nurse") passing through those associative memories will acquire a vector $v$ with gender bias information. Therefore, a natural idea would be to modify the associative memory in a way that the resulting vectors no longer carry biased information, denoted as $v^*$. In other words, the modified associative memories should satisfy:
\begin{equation}
\begin{aligned}
v^* = \hat{W}_{total}k.
\end{aligned}
\end{equation}

It's important to emphasize our aim is to maintain other effects of $W_{total}$ as much as possible while eliminating the biased information it generates for specific occupational pronouns within its associative memory. If we denote vectors representing occupational pronouns as $k_o \in E$, and vectors excluding these terms as $k_b \in P$, considering the situation involving multiple vectors, it can be approximately formulated as:

\begin{equation}\label{eq6}
\begin{aligned}
V_1 = W_{total}P,\\
V^* \approx \hat{W}_{total}E,
\\V_1 \approx \hat{W}_{total}P.
\end{aligned}
\end{equation}
$V_1$ denotes the resulting vector obtained from $k_b \in P$ after undergoing transformation via the initial associative memory $W_{total}$. Building on insights from Meng et al. \cite{meng2022mass}, we establish an optimization objective to solve Eqn. \ref{eq6} for $W_{total}$:
\begin{equation}
\begin{split}
W_{total} \triangleq \underset{\hat{W}_{total}}{\operatorname{argmin}} & \left(\left\|\hat{W}_{total} E- V^*\right\|^2 \right. \left. +\left\|\hat{W}_{total} P- V_1\right\|^2\right).
\end{split}
\end{equation}

\begin{figure}[htb]
\centering
\includegraphics[width=0.5\linewidth]{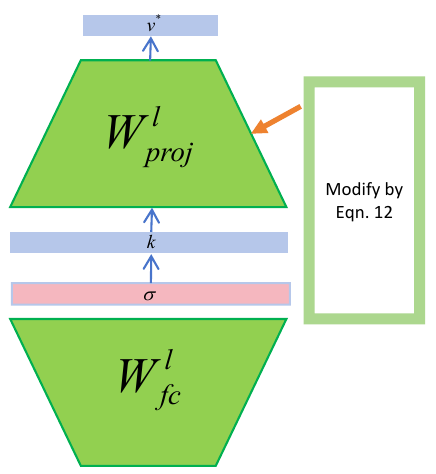}
\caption{LSDM applied in a single $W_{proj}$, modifying the parameters of $W_{proj}$ by finding the vector $k$ and the corresponding unbiased vector $v^*$.}
\label{fig5}
\end{figure}

Given the nonlinearity inherent in the definition of $W_{total}$ above, for the sake of simplicity, we opt to employ $W_{proj}$—the last linear layer within the MLP, instead of $W_{total}$. This consideration narrows our focus to a single $W$, representing a solitary linear layer\footnote{Subsequently, we'll extend our approach to encompass scenarios involving multiple linear layers.}:
\begin{equation}
\begin{split}
W \triangleq \underset{\hat{W}}{\operatorname{argmin}} \left(\left\|\hat{W} E- V^*\right\|^2 \right. \left. +\left\|\hat{W} P- V_1\right\|^2\right).
\end{split}
\end{equation}
Taking the derivative and setting it to zero gives:
\begin{equation}
\left(\hat{W}E- V^*\right)E^T+\left(\hat{W}P- V_1\right) P^T=0,
\end{equation}
and\footnote{$\hat{W}=W_0+\Delta$, this means the updated parameter $\hat{W}$ is equivalent to adding $\Delta$ to the original parameter $W_0$.}:

\begin{equation}
\left(W_0+\Delta\right)\left(E E^T+P P^T\right)=V^* E^T+ V_1 P^T.
\end{equation}
Further expansion leads to:
\begin{equation}\label{eq11}
W_0 E E^T+W_0 P P^T+\Delta E E^T+\Delta P P^T=V^* E^T+ V_1 P^T.
\end{equation}
By combining Eqn. \ref{eq6} $''WP=V_1''$ and Eqn. \ref{eq11}, we obtain:
\begin{equation}\label{eq12}
\begin{aligned}
&\Delta\left(E E^T+P P^T\right)=V^* E^T-W_0 E E^T,\\
&\Delta=(V^* E^T-W_0 E E^T)\left(E E^T+P P^T\right)^{-1}.
\end{aligned}
\end{equation}

Since pretraining remains opaque, direct access to $E$, $P$, and $V^*$ isn't feasible. Drawing from Meng's approach \cite{meng2022locating}, we estimate the uncentered covariance of $P$ via a sample of Wikipedia texts. Subsequently, we'll detail the methodology for computing $E$ and $V^*$.

\textbf{Selecting $E$ to represent occupational pronouns.} Considering the pivotal role of MLP inputs at the last occupation pronoun token (as detailed in Section \ref{causal}), we designate inputs representing the occupation pronoun at its final token as the lookup key $k^*$. To compute this, we collect activations as follows: Initially, we pass the text $x$ containing the occupation pronoun through model $G$. Then, at layer $l$ and the final occupation pronoun token index $i$, we extract the value of $W_{proj}$'s input:
\begin{equation}
k^* = \frac{1}{N} \sum_{j=1}^N k\left(s_j+x\right),
\end{equation}
where:
\begin{equation}
k(x) = \sigma\left(W_{fc}^{l} \gamma\left(a_{[x], i}^{l} + h_{[x], i}^{l-1}\right)\right) \text{.}
\end{equation}

To refine the representation of the vector $k$ to more accurately encapsulate the essence of a particular occupational pronoun, we adopt Meng's strategy \cite{meng2022mass} by adding certain prefix tokens $s_i$ to the text. This augmentation aims to bolster its resilience and subsequently compute the average. We've gathered a corpus of 5,980 texts across 299 occupational pronouns to derive the respective $k$, which collectively constitute $E$. Further elaboration on automatically generating texts with occupational bias will be provided in Section \ref{auto}.

\textbf{Constructing the bias-free vector $V^*$.} Next, we wish to construct some vector $v^*$ that encodes the non-biased information. We set:

\begin{equation}
v^*=\operatorname{argmin}_{z} \mathcal{L}(z).
\end{equation}
where the objective $\mathcal{L}(z)$ is:
\begin{equation}\label{eq16}
\frac{1}{N} \sum_{j=1}^N -\log \mathbb{P}_{G\left(\hat{m}_i^{l}:=v^*\right)}\left[o^* \mid s_{j} + x\right].
\end{equation}

$o^*$ made adjustments to the original model output to ensure that the last token output probabilities for the words ``he'' and ``she'' were $\mathbb{P}(\text{he}) = \mathbb{P}(\text{she}) = \frac{1}{2}[\mathbb{P}(\text{he}) + \mathbb{P}(\text{she})]$, while retaining unchanged prediction probabilities for other tokens. Eqn. \ref{eq16} seeks a vector $v^*$ that when substituted as the output of $W_{proj}$ at token $i$ (the last token of the occupational pronouns, denoted as $G(\hat{m}_{i}^{l}:=v^*)$), will let the network to predict the target object $o^*$ that without gender bias\footnote{As outlined in Section \ref{casual trac}, the magnitude of gender bias is denoted as $\mathbb{P}(gb) = \left|\mathbb{P}(he) - \mathbb{P}(she) \right|$.}.

To clarify, the optimization doesn't directly alter model weights but aims to get a vector $v^*$. Similar to $E$, the optimization for $v^*$ also leverages the prefix texts $s_i$ to foster robustness across varied contexts, utilizing the same 5980 texts to constitute $V^*$. Figure \ref{fig5} displays the process of modifying a single $W_{proj}$.

Once we have computed the $E$ and $V^*$, we can apply Eqn. \ref{eq12}, updating the $W_{proj}$ weights.

\textbf{Spreading over layers}. Based on the conclusions drawn in the \ref{causal}, we believe that the bottom MLP is a critical module for gender bias generation. Therefore, modifying the weight of a single $W_{proj}$ alone may not only completely eliminate the occupational gender bias within the model, but it will have a significant impact on the original performance of the model. Following Meng et al. \cite{meng2022mass}, we attempt to refine the above algorithm to achieve the ability to modify multiple $W_{proj}$.
To better illustrate the algorithm, we define:
\begin{equation}\label{eq17}
r^* = v^* - m_{i}^{l_e} = \hat{m}_{i}^{l_e} - m_{i}^{l_e}.
\end{equation}
 $l_e$ represents the last matrix that needs to be modified. $r^*$ assesses the level of divergence between vectors without gender bias and the original vectors. We aim to progressively diminish this discrepancy by adjusting parameters across multiple layers within the model, rather than concentrating solely on a single $W_{proj}$. When contemplating the necessity to modify parameters $W_{proj}$ across layers $\{l_1, l_2,...,l_e\}$, the computation incorporates:

\begin{equation}
\hat{m}_{i}^{l} = m_{i}^{l} + \frac{r^*}{l_e-l+1}.
\end{equation}
For example, if we need to modify the parameters $W_{proj}$ in layers $\{3,4,5,6,7,8\}$, we first should compute $v^*$ in layer 8 with Eqn. \ref{eq16} and $r^* = v^* - h_{i}^{8}$. Then,
in layers 6 and 8, we can get:
\begin{equation}
\begin{aligned}
\hat{m}_{i}^{6} &= m_{i}^{6} + \frac{r^*}{8-6+1}, \hat{m}_{i}^{8} = m_{i}^{8} + \frac{r^*}{8-8+1} = v^*,\\
V^{*,6} &\leftarrow\left[\hat{m}_{i_{1}}^{6}, \ldots, \hat{m}_{i_{n}}^{6}\right],
V^{*,6} \leftarrow\left[\hat{m}_{i_{1}}^{8}, \ldots, \hat{m}_{i_{n}}^{8}\right].
\end{aligned}
\end{equation}
The formula for the other layers is the same as above. On the other hand, due to the necessity of modifying multiple layers, we need to compute  $E$ and $P$ iteratively
in ascending layer order.
\subsection{Automatic Bias Sentence Generation}\label{auto}
We initially sourced the occupational dataset from Vig et al. \cite{vig2020causal}, encompassing 30 occupational pronouns for females and 269 for males. These 299 occupational pronouns encompass the vast majority of gender-biased pronouns in the corpus of large language models, and can represent occupational gender bias in large language models\footnote{\href{https://github.com/sebastianGehrmann/CausalMediationAnalysis}{https://github.com/sebastianGehrmann/CausalMediationAnalysis}}. Our approach involved utilizing the model's inherent properties to generate biased sentences. We inputted the template \texttt{`The \{occupation pronoun\}'} into the model and allowed it to autonomously complete the sentence. Setting a threshold of $d \in D$ and $f$ to generate $f$ sentences, each of length $d$, we then selected the top five hundred sentences with the highest perplexity. These sentences underwent a subsequent round: we fed them back to the model and, among the probabilities for the next word, selected five sentences with the highest $\mathbb{P}(gb)$ value. For the implementation, we chose $d \in \{8,9,10,11\}$ and $f = 6000$, yielding 20 sentences for each occupation pronoun. Algorithm \ref{alg1} summarizes the process of \textbf{LSDM}.
\begin{algorithm}
\caption{LSDM Algorithm}
\label{alg1}
\begin{algorithmic}[1] 
\REQUIRE occupation pronoun dataset $T$, threshold $D$, number of sentences $f$ to generate, layers $R$ to modify.
\ENSURE Model $G$ without gender bias
\STATE Initialize a gender bias set $X$
\FOR {$t \in T$}
    \FOR {$d \in D$}
        \STATE Generate $f$ sentences randomly;
        \STATE Compute perplexity to select $S_{top500}^{t}$;
        \STATE Selected the top 5 sentences $S_{top5}^{t} (gb)$ with the highest $\mathbb{P}(gb)$ in $S_{top500}^{t}$;
        \STATE Add $S_{top5}^{t}(gb)$ to set $X$;
    \ENDFOR
\ENDFOR
\FOR {$x \in X$}
    \STATE Optimize $v^*=\operatorname{argmin}_{z} \mathcal{L}(z)$      (Eqn.\ref{eq16});
    \STATE Compute $r^* = v^* - \hat{m}_{i}^{l_e}$ (Eqn.\ref{eq17});
\ENDFOR
\FOR {$l \in R$}
    \STATE Compute covariances $PP^T$;
    \FOR {$x \in X$}
        \STATE $k_i^l = \frac{1}{N} \sum_{j=1}^N    
                k\left(s_j+x\right)$;
        \STATE $\hat{m}_{i}^{l} = m_{i}^{l} + \frac{r^*}{l_e-l+1}$;
    \ENDFOR    
    \STATE $E^l \leftarrow\left[k_{i_{1}}^{l}, \ldots, k_{i_{5980}}^{l}\right]$;
    \STATE $V^{*, l} \leftarrow\left[\hat{m}_{i_{1}}^{l}, \ldots, \hat{m}_{i_{5980}}^{l}\right]$;
    \STATE Cpmpute $\Delta^l$ (Eqn.\ref{eq12});
    \STATE $\hat{W}^l = W^l +\Delta^l$;
\ENDFOR
\RETURN Model $G$
\end{algorithmic}
\end{algorithm}

\section{EXPERIMENTS}
\subsection{LSDM mitigates gender bias effectively}

To explore the effectiveness of different gender debias algorithms, we evaluated the effect on five large language models whose scales range from 6B to 13B parameters\footnote{All models are modified and reasoned with float16 precision on NVIDIA A100 40G GPUs.}:
\begin{itemize}
    \item[$\bullet$] \textbf{GPT-J} \cite{wang2021gpt}: An auto-regressive text generation model trained on the Pile with 6 billion parameters. As an early large language model, GPT-J has been widely used in a variety of experiments.
    \item[$\bullet$] \textbf{Llama} \cite{touvron2023llama}: A collection of foundation language models ranging from 7B to 65B parameters, trained on trillions of tokens. Llama demonstrates the ability to train state-of-the-art models using publicly available datasets exclusively, without resorting to proprietary and inaccessible datasets. We chose Llama-6B and Llama-13B as test models.
    \item[$\bullet$] \textbf{Baichuan} 1 and 2 \cite{yang2023baichuan}: The new generation of open-source large language models, trained on high-quality English and Chinese corpus. They achieved the best performance of their sizes on multiple authoritative Chinese, English, and multi-language general and domain-specific benchmarks. We chose Baichuan-7B and Baichuan2-13B as test models.
\end{itemize}

We examine four algorithms, including:
\begin{itemize}
    \item [$\bullet$]\textbf{None}, the original model without any debias algorithms.
    \item [$\bullet$]\textbf{LSDM}, as mentioned in \ref{lsdm}.
    \item [$\bullet$]\textbf{FT}, we fine-tuned the model on a dataset consisting of 5980 sentences generated from section \ref{auto}, with the training objective being:
    $-\log \mathbb{P}\left[o^* \mid x\right]$. The same as Eqn \ref{eq16}, $o^*$ made adjustments to the original model output, ensuring that the last token output probabilities for the words ``he'' and ``she'' are $\mathbb{P}(\text{he}) = \mathbb{P}(\text{she}) = \frac{1}{2}[\mathbb{P}(\text{he}) + \mathbb{P}(\text{she})]$, while keeping the prediction probabilities for other words unchanged.
    \item [$\bullet$] \textbf{CDA} \cite{zmigrod2019counterfactual}, unlike FT, $o^*$ ensured that the last token output probabilities for the words ``he'' and ``she'' are $\hat{\mathbb{P}}(he) = \mathbb{P}(she)$, $\hat{\mathbb{P}}(she) = \mathbb{P}(he)$, while keeping the prediction probabilities for other words unchanged.
\end{itemize}

For the four methods mentioned above, we adjusted parameters of the same size to ensure a fair comparison.
Three datasets were used to measure gender bias in the updated models:
\begin{itemize}
    \item [$\bullet$]\textbf{WinoGender} \cite{rudinger2018gender} is made of Winograd Schema Challenge, and biases are evaluated by determining if a model co-reference resolution performance is impacted by the gender of the pronoun. This evaluation aims to unveil societal biases linked to occupations captured by the model. For instance, \textit{'The nurse notified the patient that his shift would be ending in an hour.'} The model's co-reference resolution is examined using \textit{'her/her/she', 'his/him/he', and 'their/them/someone'} pronouns, comparing the perplexity of continuations for nurse and patient." In evaluation, we only consider the probabilities of the words "she" and "he".
    \item [$\bullet$]\textbf{WinoBias} \cite{zhao2018gender}, similar to Wingender, is used to assess gender bias in coreference resolution. The objective is to evaluate whether the coreference system exhibits gender bias by linking pronouns to gender-associated professions.
    \item [$\bullet$]\textbf{Professions Dataset}, is created by Vig \cite{vig2020causal} to analyze neuronal interventions. This dataset contains 319 occupational pronouns and 17 sentence templates. We removed 10 gender-neutral occupational pronouns, retaining 299 gender-biased occupational pronouns.
\end{itemize}

And three evaluation metrics:
\begin{itemize}
    \item [$\bullet$]$\mathbb{P}(gb) = \left|\mathbb{P}(he) - \mathbb{P}(she) \right|$, as mentioned earlier, is used to measure the gender bias of the model.
    \item [$\bullet$]$\mathbb{P}(sp) = \mathbb{P}(he) + \mathbb{P}(she)$. This metric measures the sum of the model's predicted probabilities for gender-related words, reflecting whether the model's prediction distribution for the next word attribute has been changed due to adjustments from the gender debias methods. For the model after using the debiasing algorithm, we want this value to be the same as the original model's. \textbf{In the table of experimental results, we show in bold the values that are most similar to the $\mathbb{P}(sp)$ values of the original model(None).}
    \item [$\bullet$] \textbf{Perplexity (ppl)}. For sentences within the dataset, we let the model generate subsequent sentences and measured the perplexity of these generated sentences on the original model (without using any debias algorithm). This metric reflects the degree to which the debiased model deviates from the original model. \textbf{In the experimental results table, we bold the lowest ppl. Typically, the original model achieves the lowest perplexity in its generating sentences. Therefore, we primarily compare the perplexity of models after applying different debiasing algorithms.}
\end{itemize}

\begin{table}[]
\centering
\caption{Experimental results of four debias algorithms on three bias datasets.}
\label{table1}
\resizebox{\columnwidth}{!}{%
\begin{tabular}{lllllllllll}
\hline
\multirow{2}{*}{\textbf{Model}} &
  \multirow{2}{*}{\textbf{Algorithm}} &
  \multicolumn{3}{c}{\textbf{WinoGender}} &
  \multicolumn{3}{c}{\textbf{WinoBias}} &
  \multicolumn{3}{c}{\textbf{Pro Dataset}} \\ \cline{3-11} 
 & & $\mathbb{P}(gb)$ $\downarrow$ & $\mathbb{P}(sp)-$ & $\textbf{ppl}$ $\downarrow$ & $\mathbb{P}(gb)$ $\downarrow$ & $\mathbb{P}(sp)-$ & $\textbf{ppl}$ $\downarrow$ & $\mathbb{P}(gb)$ $\downarrow$ & $\mathbb{P}(sp)-$ & $\textbf{ppl}$ $\downarrow$ \\ \hline
\multirow{4}{*}{GPT-J-6B} & None & 12.30 & 31.53 & 14.37 & 19.87 & 41.40 & \textbf{20.38} & 23.38 & 36.03 & \textbf{17.86} \\
 & FT & 7.71 & 39.80 & 18.11 & 11.18 & 46.47 & 25.22 & 10.18 & 42.99 & 22.58 \\
 & CDA & 10.27 & 42.93 & 16.35 & 12.74 & 50.72 & 23.50 & 13.60 & 47.07 & 21.27 \\
 & LSDM & \textbf{5.20} & \textbf{31.57} & \textbf{14.04} & \textbf{9.06} & \textbf{43.23} & 21.36 & \textbf{7.48} & \textbf{35.70} & 18.63 \\ \hline
\multirow{4}{*}{Llama-7B} & None & 11.25 & 26.68 & \textbf{10.76} & 19.24 & 34.19 & \textbf{14.19} & 23.96 & 32.15 & \textbf{11.98} \\
 & FT & 6.77 & 25.65 & 12.57 & 10.47 & \textbf{34.51} & 25.51 & 8.57 & 31.89 & 14.60 \\
 & CDA & 7.66 & 29.39 & 16.35 & 12.63 & 36.14 & 16.90 & 11.46 & 33.41 & 15.47 \\
 & LSDM & \textbf{5.81} & \textbf{26.32} & 11.56 & \textbf{6.25} & 32.04 & 14.92 & \textbf{5.69} & \textbf{23.43} & 12.42 \\ \hline
\multirow{4}{*}{Baichuan-7B} & None & 10.09 & 30.72 & \textbf{9.47} & 18.94 & 41.37 & \textbf{12.58} & 21.14 & 33.57 & \textbf{10.51} \\
 & FT & 8.80 & 34.87 & 11.49 & 8.54 & 44.16 & 15.80 & 14.80 & 40.94 & 13.67 \\
 & CDA & 8.70 & 29.15 & 12.32 & 12.55 & 35.83 & 14.18 & 17.77 & 30.92 & 13.33 \\
 & LSDM & \textbf{5.34} & \textbf{30.76} & 10.25 & \textbf{7.18} & \textbf{40.19} & 14.13 & \textbf{6.42} & \textbf{31.41} & 12.80 \\ \hline
\multirow{4}{*}{Llama-13B} & None & 12.96 & 34.12 & \textbf{10.13} & 19.24 & 34.72 & \textbf{13.54} & 25.50 & 35.30 & \textbf{11.15} \\
 & FT & 11.48 & 36.67 & 11.29 & 11.03 & 46.13 & 20.77 & 9.88 & 37.47 & 16.77 \\
 & CDA & 7.70 & 38.77 & 10.84 & 14.92 & 48.52 & 26.65 & 12.15 & 41.25 & 23.89 \\
 & LSDM & \textbf{6.29} & \textbf{35.57} & 10.74 & \textbf{9.64} & \textbf{35.23} & 13.71 & \textbf{9.80} & \textbf{34.88} & 11.90 \\ \hline
\multirow{4}{*}{Baichuan2-13B} & None & 11.48 & 36.67 & \textbf{8.75} & 25.53 & 45.15 & \textbf{11.79} & 27.78 & 40.56 & \textbf{10.60} \\
 & FT & \textbf{5.48} & 36.54 & 19.19 & 13.01 & 41.55 & 13.54 & 12.78 & \textbf{41.26} & 12.64 \\
 & CDA & 8.08 & 33.23 & 9.99 & 11.42 & 42.32 & 13.50 & 13.56 & 41.56 & 12.66 \\
 & LSDM & 6.07 & \textbf{36.94} & 9.56 & \textbf{9.03} & \textbf{44.66} & 12.92 & \textbf{10.58} & 39.00 & 12.69 \\ \hline
\end{tabular}
}
\end{table}

Table \ref{table1} showcases experimental results on gender bias datasets. In terms of mitigating model gender bias ($\mathbb{P}(gb)$), LSDM emerged as the most effective among the four models, achieving an average reduction of 71.4\% in gender bias. While FT showed good debias performance in WinoGender with Baichuan2-13B, its overall reduction in gender bias averaged only 40.6\%. Despite adjusting gender probabilities during training, both FT and CDA exhibited inferior debiasing results during testing, primarily due to disparities in distribution between the training and test datasets. In contrast, the LSDM method makes broad adjustments to specific gendered words without targeting entire sentences, facilitating effective debias when encountering occupation tokens during tests, without necessitating sentence-level consistency with the training data.

In terms of preserving model performance, LSDM exhibits the closest similarity to the original model in $\mathbb{P}(sp)$. This indicates that LSDM doesn't merely reduce the probabilities of ``she'' and ``he'' simultaneously to diminish gender bias $\mathbb{P}(gb)$, but dynamically adjusts their respective probabilities. Additionally, LSDM is also closest to the original model in terms of perplexity (ppl), this suggests that LSDM maximizes the preservation of the model's linguistic capabilities compared to several other debias methods. These three metrics reveal that while LSDM removes model bias, it also maximally retains the original model's characteristics.

\subsection{LSDM maintains the model's capabilities}
Another crucial aspect is that the debiased model should perform well on other datasets unrelated to gender bias \cite{guo2022auto, zmigrod2019counterfactual}. Therefore, we selected the following seven datasets that relate to knowledge, language, reasoning, and comprehension for evaluation:
\begin{itemize}
    \item \textbf{COPA} \cite{roemmele2011choice}: Evaluates open-domain commonsense causal reasoning with 1000 questions. Each question presents a premise and two alternatives, requiring the selection of the alternative with a more plausible causal relation to the premise.
    \item \textbf{RTE} \cite{wang2018glue}: A natural language inference task that determines logical relations (entailment, contradiction, neutral) between given sentence pairs.
    \item \textbf{CommonsenseQA} \cite{talmor2018commonsenseqa}: A multiple-choice question-answering dataset that necessitates various types of commonsense knowledge to predict the correct answers. It comprises 12,102 questions with one correct answer and four distractors.
    \item \textbf{HellaSwag} \cite{zellers2019hellaswag}: A challenging dataset for evaluating commonsense natural language inference, particularly challenging for state-of-the-art models but considered trivial for humans.
    \item \textbf{LAMBADA} \cite{paperno2016lambada}: Assesses computational models' text understanding capabilities through a word prediction task, measuring the model's ability to track broader discourse information.
    \item \textbf{PIQA} \cite{bisk2020piqa}: A physical interaction question-answering task that involves selecting the most reasonable solution from a given scenario and two possible solutions. Designed to test a model's knowledge of physical commonsense.
    \item \textbf{SIQA} \cite{sap2019socialiqa}: A social interaction question-answering task that requires selecting the most reasonable behavior from a given scenario and three possible subsequent behaviors. Designed to test a model's knowledge of social commonsense.
    
\end{itemize}

\begin{table}[]
\centering
\caption{Experimental results of four debias algorithms on seven model proficiency testing datasets, \textbf{$Acc$} denotes accuracy on datasets.}
\label{table2}
\resizebox{\columnwidth}{!}{%
\begin{tabular}{ccccccccc}
\hline
\multirow{2}{*}{\textbf{Model}} &
  \multirow{2}{*}{\textbf{Algorithm}} &
  \textbf{COPA}   &
  \textbf{RTE}  & 
  \textbf{commonsense}  &
  \textbf{hellaswag}  &
  \textbf{lambada}  &
  \textbf{piqa} &
  \textbf{siqa}   \\ \cline{3-9} 
                             &      & \textbf{$Acc$} $\uparrow$ & \textbf{$Acc$} $\uparrow$ & \textbf{$Acc$} $\uparrow$   & \textbf{$Acc$} $\uparrow$   & \textbf{$Acc$} $\uparrow$  & \textbf{$Acc$} $\uparrow$   & \textbf{$Acc$} $\uparrow$   \\ \hline
\multirow{4}{*}{GPT-J-6B}    & None & 68.00          & 57.76          & 56.67          & 61.81          & \textbf{69.07}          & 75.46          & 43.24          \\
                             & FT   & \textbf{71.00} & 46.93          & 42.83          & 49.87          & 42.85          & 71.44          & 39.97          \\
                             & CDA  & 70.00          & 47.29          & 42.26          & 51.82          & 46.21          & 71.82          & 40.74          \\
                             & LSDM & 68.00          & \textbf{58.48}          & \textbf{56.76} & \textbf{62.09} & 68.60          & \textbf{75.90} & \textbf{43.24} \\ \hline
\multirow{4}{*}{Llama-7B}    & None & 66.00          & 46.93          & 64.29          & \textbf{69.58} & \textbf{73.26}          & 76.82          & 44.73          \\
                             & FT   & 63.00          & 48.74          & 64.21          & 68.96          & 72.25          & 76.93          & 43.60          \\
                             & CDA  & 62.00          & 47.29          & 63.23          & 66.84          & 70.44          & 76.06          & 44.63          \\
                             & LSDM & \textbf{67.00} & \textbf{48.74} & \textbf{64.54} & 69.55          & 73.14          & \textbf{76.88} & \textbf{45.24} \\ \hline
\multirow{4}{*}{Baichuan-7B} & None & 74.00          & 53.07          & 58.89          & \textbf{64.42} & 67.16          & 77.58          & 42.02          \\
                             & FT   & 69.00          & \textbf{56.32} & 53.81          & 55.01          & 54.90          & 69.31          & 41.25          \\
                             & CDA  & 70.00          & 53.43          & 57.33          & 60.95          & 63.52          & 76.01          & 41.91          \\
                             & LSDM & \textbf{74.00} & 54.51          & \textbf{59.46} & 64.04          & \textbf{69.96} & \textbf{77.58} & \textbf{42.02} \\ \hline
\multirow{4}{*}{Llama-13B}   & None & 65.00          & 53.07          & 67.32          & 74.15          & 75.76          & 78.67          & 44.17          \\
                             & FT   & 60.00          & 54.51          & 65.93          & 72.60          & 77.86          & 77.86          & 43.96          \\
                             & CDA  & 58.00          & 50.90          & 66.50          & 70.05          & 71.01          & 77.26          & 43.40          \\
                             & LSDM & \textbf{69.00} & \textbf{55.60} & \textbf{67.65}          & \textbf{74.27} & \textbf{75.82}          & \textbf{78.94}          & \textbf{44.17} \\ \hline
\multirow{4}{*}{Baichuan2-13B} &
  None &
  71.00 &
  45.13 &
  \textbf{65.60} &
  \textbf{70.86} &
  \textbf{73.96} &
  78.13 &
  \textbf{44.47} \\
                             & FT   & 67.00          & 41.34          & 63.44          & 64.32          & 72.45          & \textbf{78.54} & 43.56          \\
                             & CDA  & 65.00          & 40.54          & 59.64          & 66.54          & 70.11          & 78.08          & 43.79          \\
                             & LSDM & \textbf{72.00} & \textbf{46.57} & 65.36          & 70.59          & 73.59          & 77.91          & 44.42          \\ \hline
\end{tabular}
}
\end{table}

Table \ref{table2} shows the results of four bias reduction algorithms on seven datasets, with LSDM showing consistently excellent performance across all models and datasets. Out of 35 evaluations, LSDM achieved optimal results in 24 evaluations. When compared to FT and CDA, LSDM only trailed FT in three cases.  It is noteworthy that LSDM outperforms the original model in several respects, possibly because when LSDM estimates the uncentered covariance of $P$ over a sample of Wikipedia texts, these texts exhibit greater consistency with the distribution in the test dataset, with fewer outlying words sampled, thereby improving the model's ability to handle similar cases. 

FT and CDA generally exhibit poor performance, as seen in the test of RTE using GPT-J-6B, with accuracies of only 46.93\% and 47.29\% respectively, far below the original model. This highlights the limitations of fine-tuning-based methods, as they tend to suffer from catastrophic forgetting to some extent. In contrast, as shown in the formula derivation in Section 3.2, LSDM achieves the optimal balance by considering both the original performance of the model ($P$) and its bias reduction capabilities ($E$).

\subsection{LSDM is proof of the casual tracing conclusion}

We chose to modify the last linear layer of the MLP module among the bottom layers $\{3, 4, 5, 6, 7, 8\}$ of the model. To validate causal tracing conclusions regarding biased information in the bottom MLP, we conducted two sets of comparison experiments: selecting middle layers $\{13, 14, 15, 16, 17, 18\}$ for GPT-J-6B and Llama-7B, $\{17, 18, 19, 20, 21, 22\}$ for Llama-13B; top layers $\{21, 22, 23, 24, 25, 26\}$ for GPT-J-6B, $\{25, 26, 27, 28, 29, 30\}$ for Llama-7B, $\{33, 34, 35, 36, 37, 38\}$ for Llama-13B.

\begin{table}[]
\centering
\caption{Experimental results of LSDM on bias datasets, where "\dag" denotes the application of LSDM at the middle layers of the model and "\ddag" denotes the application of LSDM at the top layers of the model.}
\label{table3}
\resizebox{\columnwidth}{!}{%
\begin{tabular}{ccccccccccc}
\hline
\multirow{2}{*}{\textbf{Model}} &
  \multirow{2}{*}{\textbf{Algorithm}} &
  \multicolumn{3}{c}{\textbf{WinoGender}} &
  \multicolumn{3}{c}{\textbf{WinoBias}} &
  \multicolumn{3}{c}{\textbf{Pro Dataset}} \\ \cline{3-11} 
 & & $\mathbb{P}(gb)$ $\downarrow$& $\mathbb{P}(sp)-$ & $\mathbf{ppl}$ $\downarrow$& $\mathbb{P}(gb)$ $\downarrow$& $\mathbb{P}(sp)-$ & $\mathbf{ppl}$ $\downarrow$& $\mathbb{P}(gb)$ $\downarrow$& $\mathbb{P}(sp)-$ & $\mathbf{ppl}$ $\downarrow$ \\ \hline
\multirow{3}{*}{GPT-J-6B} & LSDM & \textbf{5.20} & \textbf{31.57} & \textbf{14.04} & \textbf{9.06} & 43.23 & 21.36 & \textbf{7.48} & \textbf{35.7} & 18.63 \\
 & LSDM$\dag$ & 13.56 & 31.35 & 15.22 & 12.88 & \textbf{43.04} & 21.70 & 12.80 & 38.02 & 20.05 \\
 & LSDM$\ddag$ & 21.04 & 35.27 & 14.5 & 28.10 & 46.05 & \textbf{20.97} & 23.13 & 40.50 & \textbf{18.33} \\ \hline
\multirow{3}{*}{Llama-7B} & LSDM & \textbf{5.81} & \textbf{26.32} & \textbf{11.56} & \textbf{6.25} & \textbf{32.04} & \textbf{14.92} & \textbf{5.69} & 23.43 & \textbf{12.42} \\
 & LSDM$\dag$ & 10.30 & 25.12 & 12.24 & 10.8 & 33.65 & 16.07 & 12.31 & \textbf{32.17} & 13.08 \\
 & LSDM$\ddag$ & 15.62 & 28.76 & 14.24 & 21.39 & 36.67 & 19.33 & 18.04 & 30.26 & 18.44 \\ \hline
\multirow{3}{*}{Llama-13B} & LSDM & \textbf{6.29} & \textbf{35.57} & \textbf{10.74} & \textbf{9.64} & 45.23 & \textbf{13.71} & \textbf{9.80} & \textbf{34.88} & \textbf{11.9} \\
 & LSDM$\dag$ & 10.68 & 32.98 & 10.89 & 11.76 & \textbf{43.69} & 13.83 & 10.57 & 34.50 & \textbf{11.97} \\
 & LSDM$\ddag$ & 23.68 & 38.46 & 10.94 & 29.24 & 48.45 & 14.21 & 13.54 & 36.74 & 13.11 \\ \hline
\end{tabular}
}
\end{table}

\begin{table}[]
\centering
\caption{Experimental results of LSDM on model proficiency testing datasets, where "\dag" denotes the application of LSDM at the middle layers of the model and "\ddag" denotes the application of LSDM at the top layers of the model.}
\label{table4}
\resizebox{\columnwidth}{!}{%
\begin{tabular}{ccccccccc}
\hline
\multirow{2}{*}{\textbf{Model}} &
  \multirow{2}{*}{\textbf{Algorithm}} &
  \textbf{COPA} &
  \textbf{RTE} &
  \textbf{Commonsense} &
  \textbf{Hellaswag} &
  \textbf{Lambada} &
  \textbf{Piqa} &
  \textbf{Siqa} \\ \cline{3-9} 
 & & \textbf{$Acc$} $\uparrow$ & \textbf{$Acc$} $\uparrow$ & \textbf{$Acc$} $\uparrow$ & \textbf{$Acc$} $\uparrow$ & \textbf{$Acc$} $\uparrow$ & \textbf{$Acc$} $\uparrow$ & \textbf{$Acc$} $\uparrow$ \\ \hline
\multirow{3}{*}{GPT-J-6B} & LSDM & 68.00 & 58.48 & \textbf{56.76} & \textbf{62.09} & 68.60 & \textbf{75.90} & \textbf{43.24} \\
 & LSDM$\dag$ & 68.00 & 57.76 & 56.67 & 61.81 & \textbf{69.07} & 75.46 & 43.24 \\
 & LSDM$\ddag$ & \textbf{69.00} & \textbf{61.66} & 56.67 & 61.66 & 66.91 & 75.14 & 43.24 \\ \hline
\multirow{3}{*}{Llama-7B} & LSDM & \textbf{67.00} & \textbf{48.74} & \textbf{64.54} & \textbf{69.55} & 73.14 & \textbf{76.88} & \textbf{45.24} \\
 & LSDM$\dag$ & 65.00 & 47.29 & 63.88 & 69.19 & \textbf{73.76} & 76.82 & 44.22 \\
 & LSDM$\ddag$ & 63.00 & 47.65 & 63.80 & 69.12 & 73.28 & 76.77 & 44.68 \\ \hline
\multirow{3}{*}{Llama-13B} & LSDM & \textbf{69.00} & \textbf{55.60} & 67.65 & \textbf{74.27} & 75.82 & 78.94 & \textbf{44.17} \\
 & LSDM$\dag$ & 66.00 & 54.15 & 67.65 & 73.94 & 75.74 & \textbf{79.11} & 43.91 \\
 & LSDM$\ddag$ & 63.00 & 52.35 & \textbf{67.98} & 74.24 & \textbf{76.15} & 78.89 & 44.11 \\ \hline
\end{tabular}%
}
\end{table}

Tables \ref{table3} and \ref{table4} demonstrate the effectiveness of applying LSDM at different parts of the model. Regardless of whether it's on gender bias datasets or model proficiency testing datasets, applying LSDM at the bottom yields optimal results. Aligning with earlier causal tracing conclusions highlighting the bottom MLP as a major source of model gender bias.

\subsection{The scope of the LSDM}
In order to determine the scope of LSDM's effect, we additionally included an evaluation dataset containing 10 neutral pronouns defined by Vig et al.\cite{vig2020causal} Each neutral pronoun has 17 templates as \textbf{Professions Dataset}. These 10 neutral occupational words are not within the scope of the model modification. We hope that LSDM can also reduce bias in sentences containing these occupational words. The experimental settings are consistent with previous.

\begin{table}[]
\centering
\caption{Debias results of LSDM on neutral occupational pronouns.}
\label{table5}
\resizebox{\columnwidth}{!}{%
\begin{tabular}{cccccccc}
\hline
\multirow{2}{*}{Model} & \multirow{2}{*}{Algorithm} & \multicolumn{3}{c}{\textbf{Neutral Pronouns}} & \multicolumn{3}{c}{\textbf{Pro Dataset}} \\ \cline{3-8} 
                               & & $\mathbb{P}(gb)$ $\downarrow$ & $\mathbb{P}(sp)-$ & $\textbf{ppl}$ $\downarrow$ & $\mathbb{P}(gb)$ $\downarrow$ & $\mathbb{P}(sp)-$ & $\textbf{ppl}$ $\downarrow$ \\ \hline
\multirow{2}{*}{GPT-J-6B}      & None & 16.34 & 31.32 & 14.69 & 23.38 & 36.03 & 17.86 \\
                               & LSDM & 10.32 & 32.87 & 14.87 & 7.48  & 35.70 & 18.63 \\ \hline
\multirow{2}{*}{Llama-7B}      & None & 13.31 & 28.18 & 10.46 & 23.96 & 32.15 & 11.98 \\
                               & LSDM & 8.24  & 26.45 & 10.67 & 5.69  & 23.43 & 12.42 \\ \hline
\multirow{2}{*}{Baichuan-7B}   & None & 11.35 & 27.94 & 11.54 & 21.14 & 33.57 & 10.51 \\
                               & LSDM & 7.88  & 26.98 & 11.67 & 6.42  & 31.41 & 12.80 \\ \hline
\multirow{2}{*}{Llama-13B}     & None & 15.54 & 26.96 & 10.16 & 25.50 & 35.30 & 11.15 \\
                               & LSDM & 10.12 & 27.21 & 11.03 & 9.80  & 34.88 & 11.90 \\ \hline
\multirow{2}{*}{Baichuan2-13B} & None & 18.43 & 35.64 & 8.43  & 27.78 & 40.56 & 10.60 \\
                               & LSDM & 12.23 & 33.41 & 8.55  & 10.58 & 39.00 & 12.69 \\ \hline
\end{tabular}%
}
\end{table}

Table \ref{table5} compares the debias results of sentences containing occupational pronouns after applying LSDM. The ``Neutral Pronouns" contains sentences with neutral occupational pronouns, while the ``Pro Dataset" category contains sentences with clearly gender-biased occupational pronouns. It can be observed that despite being labeled as neutral occupational pronouns, they still exhibit some gender bias, indicating the pervasive nature of gender bias and the complexity of completely eliminating it.

Although sentences containing neutral occupational pronouns were not used in LSDM, it still managed to reduce gender bias by 33\% in ``Neutral Pronouns''. In terms of $\mathbb{P}(sp)$ and \textbf{$ppl$}, LSDM remains close to the original model. This suggests that LSDM is still able to debias some unseen gender occupational pronouns bias to some extent while maintaining the model's performance.

\section{Conclusion}

This article investigates the storage locations and generation mechanisms of gender bias in large language models and proposes a gender debias method called LSDM based on knowledge editing. We employed a causal tracing method to pinpoint the location of model bias generation and suggest that gender bias is generated due to the information enrichment effect of the bottom MLP modules and is extracted by top attention modules to influence model output. Based on this, we use the least squares algorithm to modify the bottom MLP module to eliminate model bias. The experimental results show that LSDM is an efficient debias method, which not only successfully removes most of the occupational gender bias in the sentences, but also overcomes the problem of catastrophic forgetting that exists in all the traditional methods, and LSDM can be extended to occupational pronouns that have not been seen before.

In our future work, we aim to delve deeper into the mechanics of this debias method to thoroughly understand and apply it to enhance the model's knowledge performance.

\section{Limitations}
This paper conducts an in-depth analysis of the causes of gender bias and proposes an effective LSDM method to alleviate this challenge. Despite the significant performance of the proposed method, there are still some limitations. In terms of causal tracing, although most components have less impact on the final outcome, how these components collectively influence the model output has not been extensively studied in this paper. While LSDM achieves good bias reduction effects on some datasets by targeting specific occupational pronouns, its effectiveness remains poor for occupational pronouns outside the distribution. Further, this paper focuses only on bias between gender and occupational pronouns, how to extend this method to a broader range of biases remains to be explored in future research.
\bibliographystyle{splncs04}
\bibliography{refs}
\end{document}